\documentclass[letterpaper, 10 pt, conference]{ieeeconf}
\IEEEoverridecommandlockouts  
\usepackage[utf8]{inputenc}
\usepackage{graphicx}
\graphicspath{{./images/}}
\usepackage{multicol,color}
\newcommand{\RNum}[1]{\uppercase\expandafter{\romannumeral #1\relax}}
\usepackage{slashbox}
\usepackage[ruled,vlined]{algorithm2e}
\usepackage{amsmath,amsfonts,amssymb}
\newcommand{\camel}{{\texttt{CAMEL}}}
\usepackage{subfig}
\title{\LARGE \bf CAMEL: Learning Cost-maps Made Easy for Off-road Driving}
\author{Kasi Viswanath, P.B. Sujit and Srikanth Saripalli%
\thanks{Kasi Viswanath and P.B. Sujit are with IISER Bhopal, Bhopal-- India. e-mail:(kasi18,sujit)@iiserb.ac.in}
\thanks{Srikanth Saripalli is with the Department of Mechanical Engineering, Texas A\&M University, College Station, Texas, TX--  77843-3123. e-mail:{ssaripalli@tamu.edu}}}
\begin{document}
\maketitle
\thispagestyle{empty}
\pagestyle{empty}
\begin{abstract}
Cost-maps are used by robotic vehicles to plan collision-free paths. The cost associated with each cell in the map  represents the sensed environment information which is often determined manually after several trial-and-error efforts. In off-road environments, due to the presence of several types of features, it is challenging to handcraft the cost values associated with each feature. Moreover, different handcrafted cost values can lead to different paths for the same environment which is not desirable. In this paper, we address the problem  of learning the cost-map values from the sensed environment for robust vehicle path planning. We propose a novel framework called as \camel~ using deep learning approach that learns the parameters through demonstrations yielding an adaptive and robust cost-map for path planning. \camel ~has been trained on multi-modal datasets such as RELLIS-3D. The evaluation of \camel ~is carried out on an off-road scene simulator (MAVS) and on field data from IISER-B campus. We also perform real-world implementation of \camel ~on a ground rover. The results shows flexible and robust motion of the vehicle without collisions in unstructured terrains.
\end{abstract}

\section{Introduction}
Unmanned ground vehicles (UGV) are used in several terrains and off-road conditions for applications such as search and rescue, surveillance, inspection, exploration etc. In these unstructured environments with varying texture and slopes, achieving autonomous capability through planning is more difficult than structured urban environments. For autonomous traversal, cost function and maps are used to encapsulate the terrain features from the perception module. Many planning systems employ manually designed cost-maps and cost-functions \cite{eb67dba83be74fb385808141788f7602} with successful demonstrations at the DARPA Grand Challenge \cite{wiley2006}. Generally, these cost functions have obstacles inflated with respect to the vehicle size. The general weighting of costs to different perception sensors are handcrafted through numerous trials which is a laborious, time consuming and relies on detailed domain knowledge. Moreover, with different weighting can lead to the generation of different paths which is not desired. 

\begin{figure}\centering
    \subfloat[]{\label{fig:scene}\includegraphics[width = 5 cm,height=4cm]{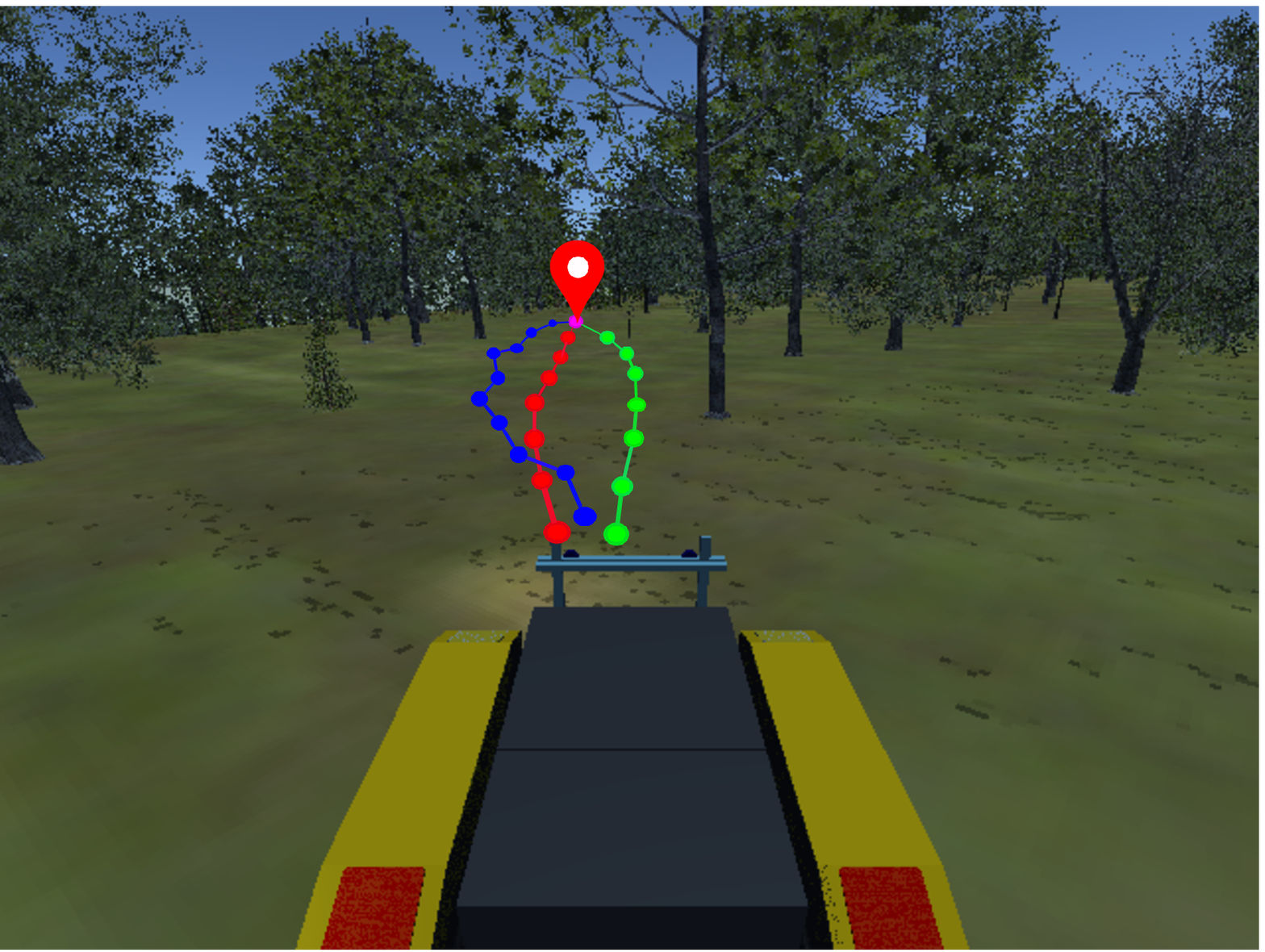}}
    \subfloat[]{\label{fig:vehicle}\includegraphics[width = 3.3 cm,height=4cm]{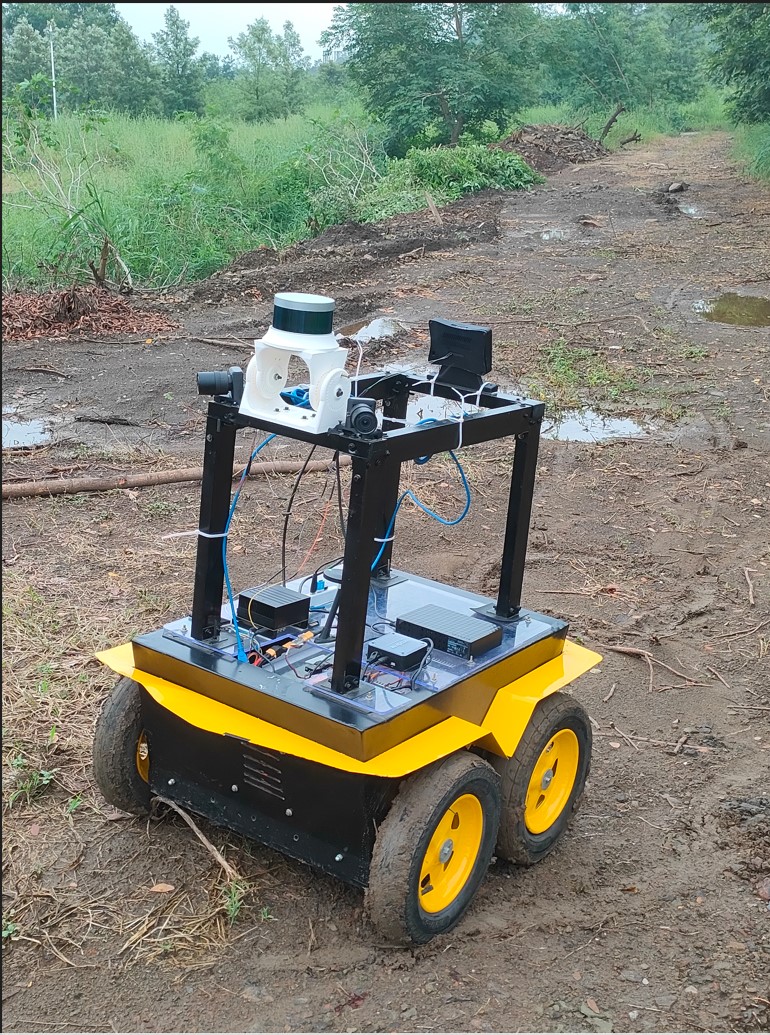}}    
\caption{(a) Trajectories generated while navigating in a simulated off-road environment. \textcolor{red}{Trajectory generated by \camel}; \textcolor{green}{Trajectory generated from a finely handcrafted cost-map}; \textcolor{blue}{Trajectory generated from coarsely designed cost-map.} (b) UGV used for real-world experiments. }
\vspace{-0.4cm}
\end{figure}

Consider a surveillance application where a robot needs to navigate in a forest area (as shown in Figure \ref{fig:scene}), the cost-map needs to estimate traversable regions coping with obstacles, trees, bushes, marshy area, puddle and irregular elevations. These scenarios introduce new challenges to the planning module and determining an optimal cost-map parameters through handcrafting would prove to be an inefficient strategy. In Figure \ref{fig:scene}, we can see that the paths for the handcrafted costs is not optimal. This issue, motivates us to ask the question: Can we learn a cost-map taking the camera and LiDAR information directly such that the cost-map represents information similar to the expert human demonstration?

In this paper, we present an approach called \camel~\footnote{As camels can walk in different terrains, our proposed framework is applicable to different types of terrains and hence we named our architecture CAMEL.}  that uses fully convolutional networks (FCN) to fit the multi-modal data and generate cost-maps through expert demonstrations. Deep Learning architecture enables learning high versatile, highly non-linear models necessary for complex and dynamic environments. Although, learning cost maps has been a topic of interest in the robotics community, however there are very few articles in this topic. In \cite{9812238} and \cite{doi:10.1177/0278364917722396}, deep reinforcement learning and deep inverse reinforcement learning techniques respectively are used to learn the cost-maps. 
However, in these approaches, initially a pre-trained CNN model trained on manually designed cost-map is used. The cost-map is further refined by the RL framework. This approach confines the model to the manually designed scene conditions. Due to the varying off-road scene conditions, this approach requires extensive amount of demonstration data to capture the complexity of off-road environment. In \camel, we bypass a pretrained model by directly learning to process the visual (camera) and point cloud (LiDAR) data from the perception module to generate trajectories that mimic human driving.
\begin{figure*}
    \centering
    \includegraphics[width = \textwidth, height = 10 cm]{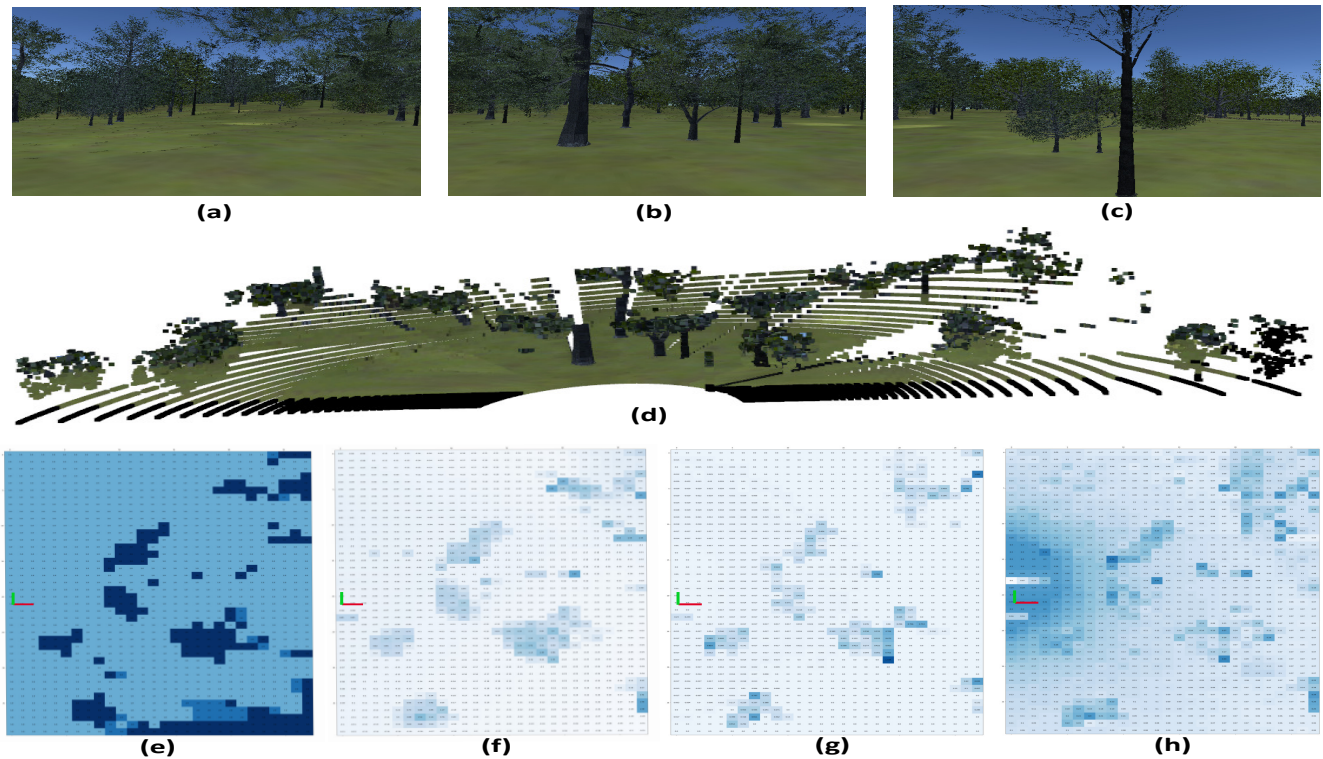}
    \caption{Figure shows the data generated from a single instance in simulation. (a), (b) and (c) are the images from left, forward and right camera respectively. (d) The three images projected to point cloud data. (e) The map of projected semantic segmentation on point cloud. The colours range indicates class hierarchy where blue represents traversable region and dark blue to obstacles. (f) The average height map extracted from the point cloud data. (g) The estimated slope map indicating the elevation of the terrain. (h) The LiDAR reflectance intensity map.}
    \label{fig:data}
    \vspace{-0.3cm}
\end{figure*}

While traversing in off-road terrains, the vehicle stability and safety is essential and hence the  visual and geometric features of the terrain such as vegetation, slope, soil stability need to be considered to plan robust and adaptive paths for the vehicle. These features are obtained by the sensors like camera, LiDAR and radar. Multi-modal datasets \cite{9561251} \cite{valada16iser} \cite{maturana2018real} provides reliable images and point cloud data of diverse scenes to develop algorithms for the outdoor domain. The \camel ~framework uses semantic projections, height map, slope and LiDAR reflectance intensity to generate a cost-map yielding the trajectories which is then compared to the actions of a human driver. 


The main contributions of this paper are:
\begin{itemize}
    \item A FCN based Deep Learning framework \camel~ able to predict 2D navigation cost-map for off-road terrains. 
\item  A navigation stack module optimized to  the generated cost-maps with the ability to handle intricate and dynamic topography. \camel~ being learned through human demonstrations, alongside the navigation stack generate trajectories that mimic human driving.

\item Demonstration of the framework's scalability from simulation training to real-world implementation (Sim2Real) with very few tuning parameters.

\item Demonstration of the framework's robustness in handling sensor miscalibration and system biases in the real-world experiments. 

\item We compare the predicted cost-map against a carefully handcrafted cost-map to show the efficacy of \camel.
\end{itemize}
\section{\camel~ methodology and architecture}
The \camel ~methodology consists of three stages: input data generation,  model training and a navigation stack as shown in Figure \ref{fig:fullstack}. 
\subsection{Input data generation}
The input to the model consists of 4 different information coming from 3 cameras and a LiDAR as shown in Figure \ref{fig:data}(a)-(d). These are (i)  semantic projection (ii) average height map (iii) slope estimation  and (iv) LiDAR reflectance intensity. We fuse all these four geometric and semantic information into a grid map of prescribed size and grid resolution. The dimensions and resolution of the grid map is decided based on the sensor ranges and the vehicle dimension. Assume the cost map size is  $\ell \times w$, where $\ell$ is the length of the map and $w$ is the width of the map and the map is discretized with resolution $\Delta$. Each cell in the map is represented as $\gamma$. The vehicle's position on the map is considered as the origin. The visual data and LiDAR data are fused to each of the cells in the grid. We now describe the generation of different inputs. 
\subsubsection{Semantic projection}
For semantic segmentation we use Offseg\cite{9551643} a semantic segmentation framework for unstructured terrains which generates a 2D mask for each pixels in the input RGB image classifying into different classes. The index of the classes is assigned based on an ascending risk factor (i.e, traversable to obstacle).
The mask is then projected to the corresponding timestamped point cloud using calibration matrices for respective cameras. The projected points are then mapped onto the grid where for each cell value, the highest class index encompassed within the cell is assigned. Figure \ref{fig:data}(e) is the resultant output using the camera input from Figure \ref{fig:data}(a)-(c) and lidar input from Figure \ref{fig:data}(d). 
\subsubsection{Geometric Characteristics}
The LiDAR data provides spatial information as a point cloud. The sensor emits laser pulses which reflects off a surfaces of vegetation, buildings etc. The reflections are captured and in turn processed to provide the position of the surface in coordinate space. We compute different morphological and geometrical information of voxels generated from the point cloud.
\paragraph{{Average Height map}}
The information on the height profile of a terrain is necessary to safely navigate through the environment. This is extracted from the point cloud data. If the value of the cell is high implies there is an obstacle in the cell and if the value is low then is implies that its a safe cell to navigate as there is no obstacle in the cell. 

To generate the height map, the average height of each cell $H_\gamma$ is computed using voxels. Assuming $n$ number of voxels in $\gamma$, $H_\gamma$ is computed as the average height value of the voxels corresponding to cell $\gamma$. It is computes as
\begin{equation}
    H_\gamma =  \frac{1}{n} \sum_{\nu =1}^n H_\nu
\end{equation}
Figure \ref{fig:data}(f) shows the average height maps for an instance based on the informaion from Figure \ref{fig:data}(d). 
\paragraph{{Slope Estimation}}
By generating voxels, we take the neighbouring points in a cubic volume and do an averaging thereby giving us better estimates. Slope $\lambda$ of a voxel is the angle between its surface normal $\eta$ and the z-axis of world coordinate. It is computed by taking the cosine inverse of surface normal component to the z-axis of world coordinate. The slopes of the voxels are then averaged corresponding to their grid cells as 
\begin{equation}
    \lambda_\gamma = \frac{1}{n} \sum_{\nu =1}^n arcos(\eta_\nu^z).
\end{equation}
\begin{figure*}
   \centering
    \subfloat[]{\label{fig:fullstack}\includegraphics[width = 6.5 cm,height = 3.5 cm]{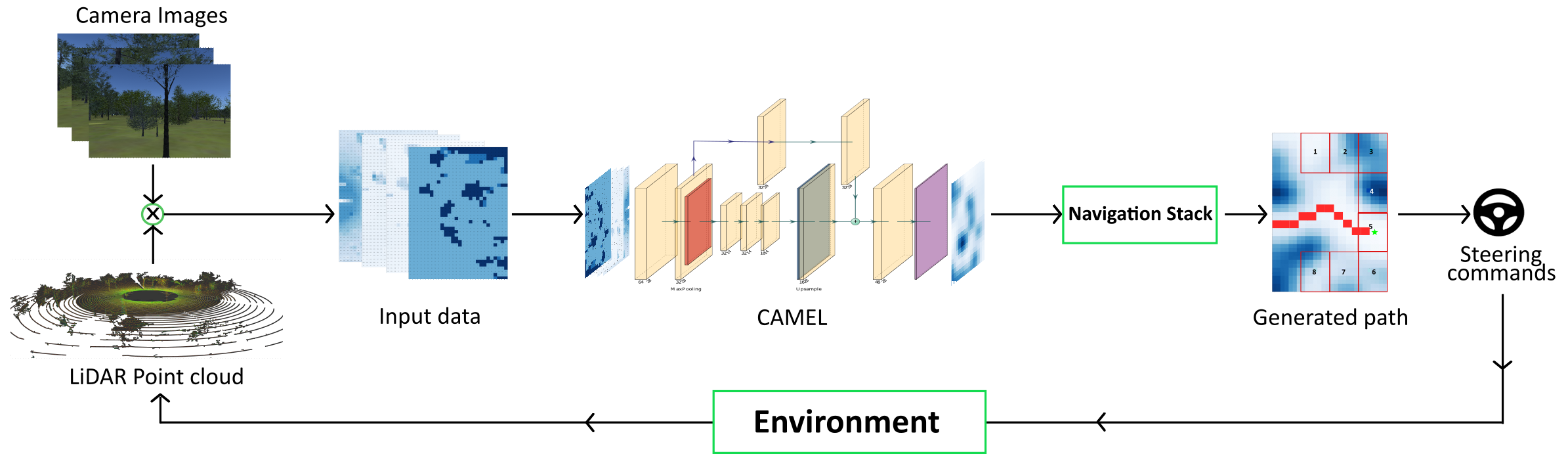}}
     \subfloat[]{\label{fig:arch}\includegraphics[width = 6.5 cm,height = 4 cm]{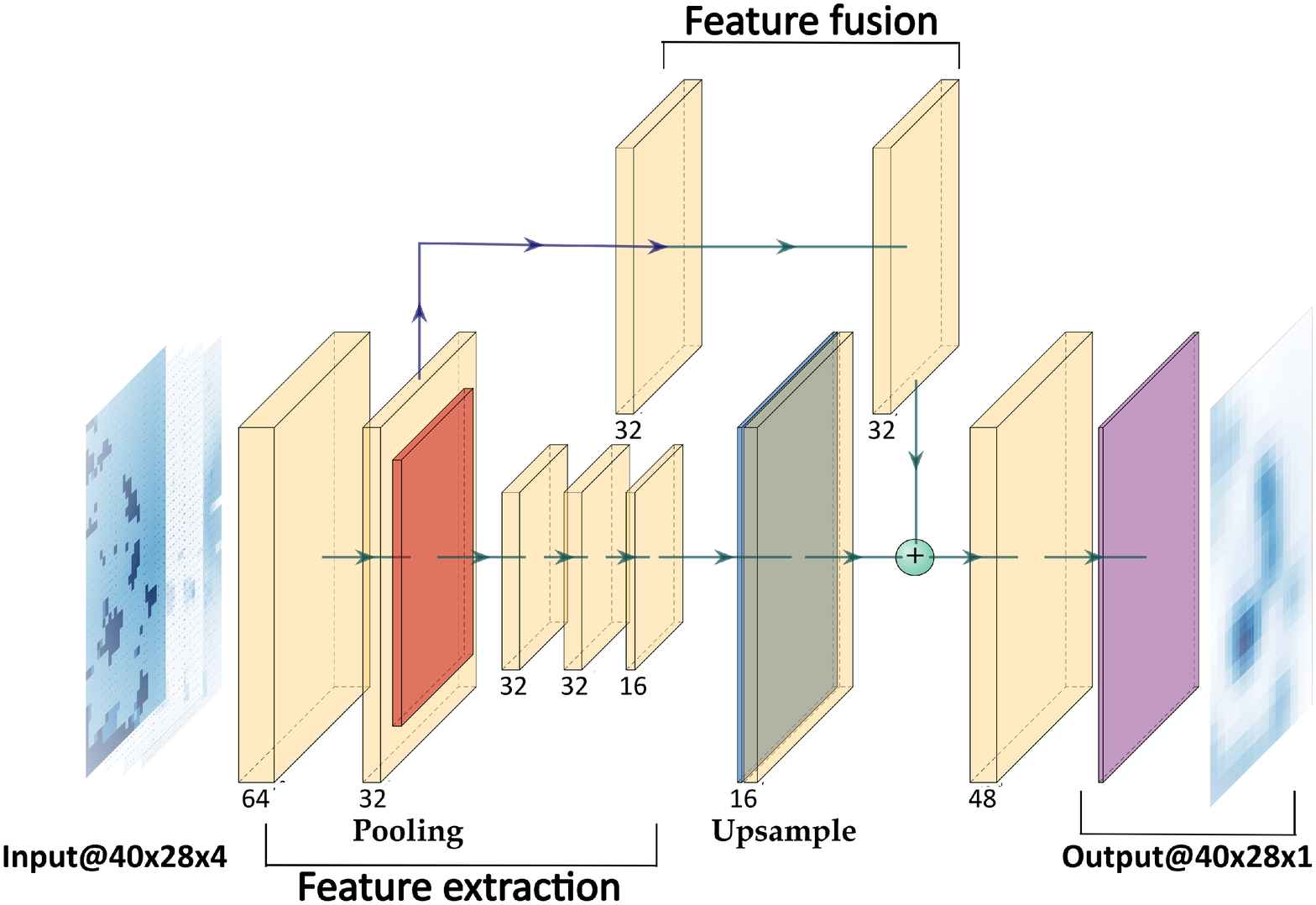}}
     \subfloat[]{ \label{fig:kernel}   \includegraphics[width = 5 cm,height = 4 cm]{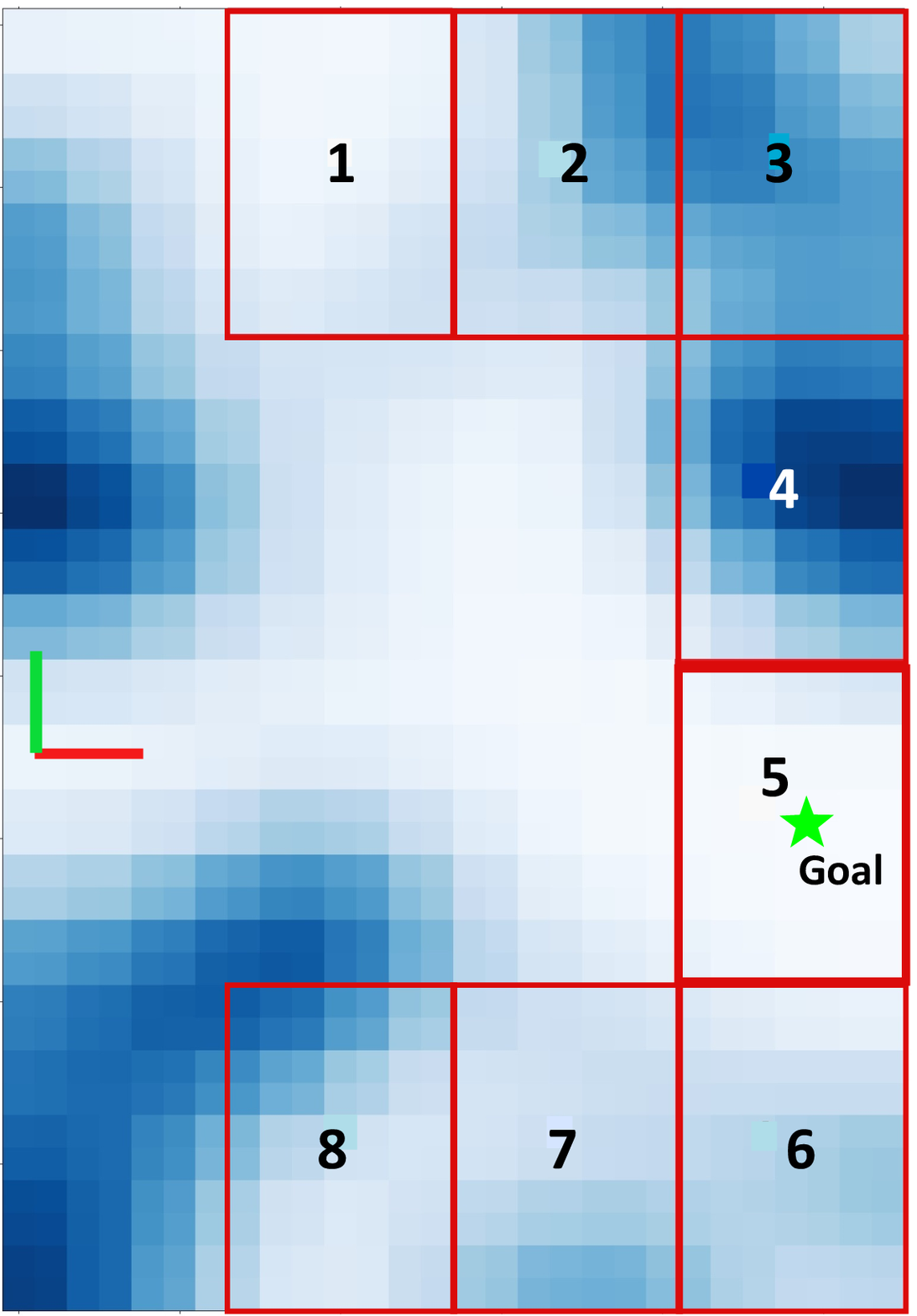}}

    \caption{(a) The complete system framework consisting of data input, \camel, and the navigation stack (b) The proposed \camel ~architecture with feature extraction and feature fusion branches (c) The kernel spaces used to find the local goal.}
    \label{fig:arch}
    \vspace{-0.3cm}
\end{figure*}

For better slope estimates $k$-dimensional tree ($k-d$ tree) is used to search the nearest neighbours. The radius for k-d tree is two times the voxel size with a maximum of 10 neighbours for better estimates. Figure \ref{fig:data}(g) shows the slope map for the input given in Figure \ref{fig:data}(d).
\paragraph{{LiDAR reflectance intensity}}
LiDAR intensity $\iota$ can be used to classify terrains and vegetation \cite{7354098}. The intensity by which the laser pulses are reflected depends on different morphological properties such as moisture content, roughness, range and surface composition. These characteristics could provide distinctive features on classes such as puddle, grass, asphalt apart from semantic segmentation. For each grid cell we compute the average intensity of the voxels as
\begin{equation}
    \iota_\gamma =  \frac{1}{n} \sum_{\nu =1}^n \iota_\nu.
    \label{iota}
\end{equation}
Figure \ref{fig:data}(h) shows the slope map for the input given in Figure \ref{fig:data}(d).

\subsection{\camel ~Architecture}
We develop a novel architecture based on Fully Convolutional Network called \camel ~to generate cost-map through demonstrations based on the perceived data from sensors. This could be considered as an imitation learning approach wherein a model tries to imitate decisions taken by human experts at a given instance. Inspired from Multi-Scale Fully Convolutional Networks\cite{doi:10.1177/0278364917722396}, the architecture consists of three segments: a feature extraction, a feature fusion as shown in Figure \ref{fig:arch} along with a navigation module to extract the steering values from the output cost-map.

For the feature extraction module, we employ four layers of convolutions and a pooling layer as listed in Table \ref{tab:arch_table} to ensure the low-level feature sharing is valid. All convolutional layers are standard Conv2D with  max-pooling layer. The convolutional layers employ a single stride, replicate padding followed by leaky-ReLU activation\cite{Maas2013RectifierNI}. The first layer has spatial kernel size of 5 $\times$ 5 and the other three with 3 $\times$ 3 kernel size.
\begin{table}
\centering\resizebox{0.45\textwidth}{!}{%
\begin{tabular}{c c c c c c c}
    \hline
    &Input & Block & c & k & s & Output\\
    \hline
    &$40\times 28\times 4$ &  Conv2D & 64  & 5 & 1 & $40\times 28\times 64$\\
    &output1 &  Conv2D & 32 & 3 & 1 & $40\times 28\times 32$\\
 feature    &output2 &  MaxPool2D & 32 & - & - & $20\times 14\times 32$\\
 extraction   &output3 &  Conv2D & 32 & 3 & 1 & $20\times 14\times 32$\\
    &output4 &  Conv2D & 32 & 1 & 1 & $20\times 14\times 32$\\
    &output5 &  Conv2D & 16 & 1 & 1 & $20\times 14\times 16$\\
    \hline
    &output6 &  Upsample & 32 & - & - & $40\times 28\times 16$\\
feature    &output2 &  Conv2D & 32 & 3 & 1 & $40\times 28\times 32$\\
fusion     &output7 &  Conv2D & 32 & 1 & 1 & $40\times 28\times 32$\\
    &output8+output6 & Concatenate & 64 & - & - & $40\times 28\times 48$\\
    &output8 & Conv2D & 1 & 1 & 1 & $40\times 28\times 1$\\
    \hline
\end{tabular}}
\caption{\camel ~uses standard convolutions (Conv2D) with two branches, a feature extractor and a feature fusion. The parameters c, k, s represents number of output channels, kernel size and stride parameters.}
\label{tab:arch_table}
\vspace{-0.5cm}
\end{table}

A skip connection is introduced which takes the output of the second Conv2D layer before max-pooling and computes two convolutions with kernel seizes 3 $\times$ 3 and 1 $\times$ 1 following leaky-ReLU activation. This enables the model to preserve translational variant and invariant features. The output from feature extraction module is then upsampled by a factor of 2 which is concatenated with the skip connection. This enables the model to treat feature channels separately. A final convolution is applied on the concatenated output yielding the cost-map.

Using steering value $y\epsilon[-1,1]$ as the ground truth, a path from start to a desired goal is generated from the output cost-map yielding steering value $\hat{y}$ which is used to compute the training loss. Mean squared error (MSE) loss function gives the loss between the target and predicted steering values  backpropagated for weight updation. Adam optimizer\cite{kingma2014method} along with weight decay ($L2$ regularization) to avoid the exploding gradient of weights. The extraction of steering value from cost-map is discussed broadly in the next section.
\begin{equation}
    MSE = \frac{1}{n}\sum_{i = 1}^n (y_i - \hat{y}_i).
\end{equation}
\subsection{Navigation Stack}\label{sec:nav}
Along with the cost-map predictor we propose a navigation module compatible and optimized for the generated cost-map consisting of few parameters. The input data $I \epsilon R^{4\times l\times w}$ is passed to the model giving an output cost-map. The cost-map $C$ is then normalized to 1 i.e, $C \epsilon [0,1]$ for easier parametrization.

For navigation a path needs to be generated in $C$ that has the least cost. This task is achieved by using a path planning algorithm like $A^*$\cite{Hart1968} or Dijkstra\cite{dijkstra1959note} for a given start and goal. The start location is the current position of vehicle, however we need to provide a goal in $C$. Here we introduce a framework for selecting a favourable goal for local path planning from $C$. To compute the local goal, a kernel space of \textcolor{black}{one sixteenth the map} dimension is considered at different locations in $C$ as shown in Figure \ref{fig:kernel}. For all the kernels, the mean and the least cost cell value within the kernel is computed and summed giving a \textcolor{black}{traversability} coefficient $T$ as given in equation \eqref{eq:trav_sc}. Now based on the current orientation of the vehicle towards goal $g$ in the global frame, a finely tuned weight  is generated for each kernel where the kernel $k$ oriented in the direction of $g$ has least weight. Weights are then distributed across other kernels based on the euclidean distance from $k$. These weights are then multiplied with $T$ to give the final kernel \textcolor{black}{risk} coefficients $V$. 
\begin{equation}
    T_k = \frac{1}{m} \sum_{i = 1}^m x_i + \min(x_1,x_2,....x_m)
    \label{eq:trav_sc}
\end{equation}
where $m$ is the number of cells in kernel k.

Considering the distance between two adjacent kernels as unit measure. Weights for kernel $i$ can be calculated from:
\begin{equation}
    W_i = W_k\times dist(k,i)
\end{equation}
where $W_k$ is the tuned weight of the kernel oriented towards $g$.
The risk coefficients $V$ are generated by multiplying traversability scores with the corresponding weights and the kernel. The kernel with least $V$ is selected as the goal kernel $k$ with the cell having least cost in the kernel as local goal point.
\begin{equation}
    V_i = W_i\times T_i
\end{equation}
for kernel $k_i$ and
\begin{equation}
    k = argmin(V)
\end{equation}
where kernel $k$ is the local goal kernel.
Now with the start and goal we compute a path planning algorithm to find least cost path for traversing. From this path a series of steering and throttle values are generated for the vehicle to reach the goal. 
By employing the navigation stack, an optimal goal in the cost-map is found considering the occupancy of neighbouring cells, cost value of the goal, orientation with respect to the global goal thereby generating a low risk, least cost path. Further discussion regarding different parameters, kernel sizes, weights are described  in the next section.

\section{Experiments}
The \camel ~framework is trained and tested on both simulation and real world scenarios. The simulations are done on Mississippi State University Autonomous Vehicle Simulator (MAVS) \cite{7988748} for different scenes while for real world, we trained the algorithm on RELLIS-3D\cite{9561251} dataset and tested on data collected in the IISERB-campus. Further, we have \camel ~implemented on a real robot. This section includes the experiments performed in the simulation framework followed by real-world implementation.
\subsubsection{Simulation}
Initial testing of the framework was performed on simulation which includes data collection of scenes, parameter tuning and navigation testing. The vehicle used is an inbuilt skid-steer model of Clearpath Robotics Warthog UGV with sensors mounted and integrated to ROS. In this section we will discuss on sensor setup, data collection, model training and navigation.

\textbf{Sensor setup:}
We use three cameras with similar intrinsic parameters and a Velodyne HDL-64E LiDAR where the three cameras are positioned and oriented to give a $180^{\circ}$ overlap field of view as given in Figures \ref{fig:data}. The vehicle has a GPS, an imu sensor and an odometry sensor which provides position and orientation of the vehicle in world frame.

\textbf{Data collection:} For training \camel, we collect the sensor data while the vehicle is driven by a human expert. The driver inputs throttle$\epsilon[0,1]$ and steering value$\epsilon[-1,1]$ which is updated to the vehicles controls through teleoperation. The data was collected for three different scenes then processed to remove desynchronized data and outliers from the pool. This throttle and steering value is used as ground truth to train the model.

We constructed a grid of dimension $40\times 28$ with the grid resolution being $0.3m\times 0.3m$. The grid resolution is taken approximately one third the vehicle width for a stable navigation. With these parameters we cover an area of 112.8 $m^2$ in front of the vehicle. The point cloud belonging to this region is extracted and converted to voxels with voxel size of 0.15$m$ (half of grid resolution). This gives a good estimate for point localization and surface normal estimates. These voxels are used to extract the geometric data grids such as average height map, slope estimation, LiDAR reflectance intensity map.  

The semantic segmentation masks for the three cameras were generated through Offseg pretrained on RELLIS-3D dataset and the indices arranged in a risk based ascending order. Offseg gives an average mIoU of 78$\%$ on the three scenes which ensures the segmentation data used for semantic mapping on LiDAR points are true. Offseg gives predictions in four classes namely sky, traversable, non-traversable, obstacle. These masks are then projected to the LiDAR voxels for generating the semantic grid map. \textcolor{black}{The approaches \cite{9812238} and \cite{DBLP:journals/corr/abs-2109-06250} uses the semantic information only to coarsely differentiate the grid cells between traversable and non-traversable. Here we provide a grouping of grid cells based on semantics and hierarchy of classes (dependent on traversability) along with geometric information to yield better cost estimates by \camel}.   

These four grid maps of an instance is appended together to create a $R^{40\times28\times4}$ input data for the model to train. A total of 4000 data instances from the simulation were generated which was shuffled and divided between train, validation and test dataset in the ratio $7:2:1$. For grid cells encompassing zero voxels, we perform a neighbourhood interpolation. This is because $nan$ values in the data leads to \textcolor{black}{huge loss of information while training \camel}.  

\textbf{Model training:} We conducted experiments of \camel ~on PyTorch framework using Python. The workstation used has a configuration of Nvidia RTX 3060 with CUDA 11.1 and CuDNN v8.1. We used 100 instances to report the average frames per second (fps) measurement.

ADAM optimizer with a learning rate of $10^{-4}$ and sequentially increasing weight decay is used to train the model for 300 epochs. Here the output of the model being a cost-map of dimension $40\times 28\times 1$ and the ground truth being a steering value, we need to extract a steering value from the cost-map to compute the loss. We employ the navigation stack on tensors in order to maintain the gradients and extract the steering value. The MSE loss is then computed between the target and predicted steering values which is backpropagated to the model for weights updation. 

\begin{figure}[t]
    \centering
    \includegraphics[width = 8cm]{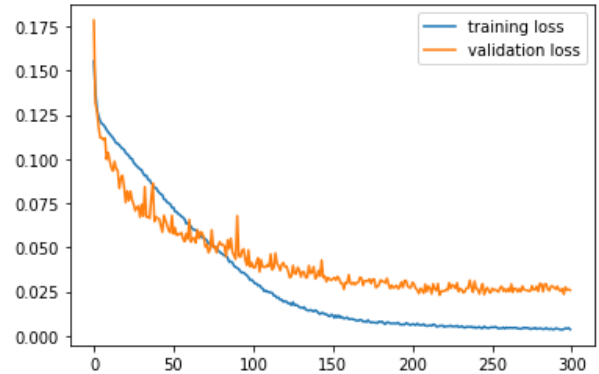}
    \caption{Training curves on simulation data. Loss over epochs for training and validation datasets are shown. The least validation loss of 0.0256 is achieved at the $217^{th}$ epoch.}
    \label{val_loss}
    \vspace{-0.4cm}
\end{figure}

The training and validation loss are shown in figure \ref{val_loss} where the model attains convergence with the least loss of 0.0256 on validation dataset whereas the test dataset gave a loss of \textcolor{black}{0.0213}. The best model is then used to test on real time simulation where the vehicle explored the environment without collisions.

\textbf{Navigation:} The best epoch from training is used to test \camel ~in the simulation. For every instance the data was processed and  used to predict the cost-map for the vehicle to navigate. This cost-map is then processed by the navigation stack which extracts the throttle and steering commands. For a smooth and jerk free motion of the robot, throttle and steering values are limited at 0.7 and [-0.4, 0.4] respectively. Parameters such as weights for goal kernel and threshold for obstacle kernel were tuned through trials to get a robust motion of the vehicle. The threshold for obstacle kernel was found optimal at 0.625 for simulations. 

While training we considered the costs of three adjacent grid cells to compute the least cost-map using Dijkstra thereby incorporating the size of the vehicle for steering clear from obstacle at safe distances. Figure\ref{fig:scene} shows trajectories generated from \camel ~and handcrafted cost-maps. While testing we observed that the path generated by Dijkstra considering single grid cell cost is identical to the one generated by considering costs from three grid cells. This suggests that the vehicle specific characteristics also influences \camel ~in learning the cost-map. 

\begin{figure*}[!h]
    \centering
    \includegraphics[width =\textwidth, height = 4.5 cm]{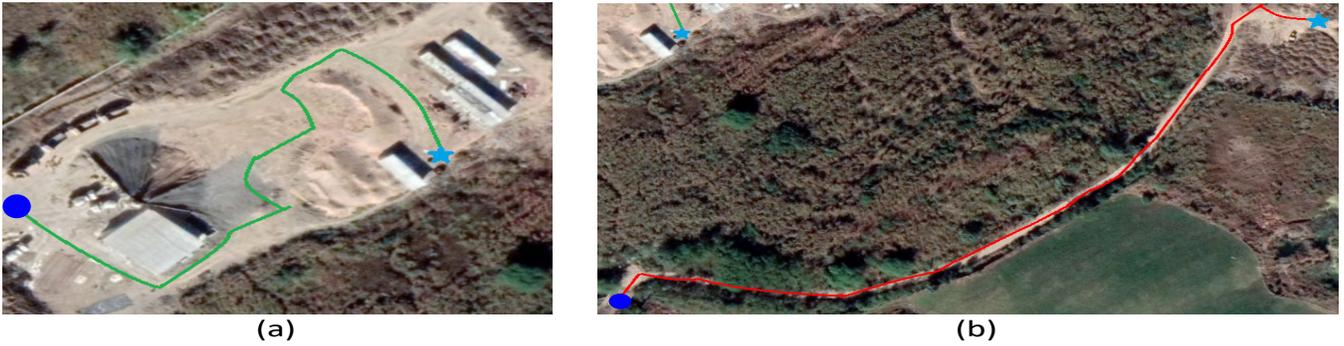}
    \caption{GPS tracks of the vehicle while traversing autonomously in IISER-B campus during testing at two different regions. The terrain included steep climbs, puddles, dynamic obstacles where the vehicle performed robust planning. Blue circle $\rightarrow$ starting location, blue star$\rightarrow$ mission end. The vehicle traversed a total distance of 450 meters in the two tracks.}
    \label{fig:tracks}\vspace{-0.2cm}
\end{figure*}
\subsubsection{Real World scenarios}The models trained on simulation is used for testing at various locations of IISER-B campus. We use a skid-steer vehicle with comparable dimensions to the Clearpath Robotics Jackal UGV mounted with calibrated sensors and Jetson AGX Xavier platform for on-board computations.

\textbf{Sensor setup:} We use two Teledyne Flir Firefly S mounted at specific angles giving a $180^{\circ}$ field of view overlap along with Velodyne VLP-16 LiDAR both calibrated. The vehicle uses PixCube Orange for flight controls consisting of GPS, compass and IMU sensor giving us the position and orientation of the vehicle. 

\textbf{Navigation:} The best \camel ~model obtained from simulation was used for navigation in real-world. Offseg pre-trained on RUGD dataset\cite{8968283} generated the semantic segmentation as it is found to produce acceptable results in IISER-B campus scenes. The computation time taken to process a single instance on average is 0.95 seconds. The output throttle and steering from the current instance is executed for 1 second as the maximum velocity of the vehicle is 0.5 m/s where the commands generated are valid for up to 1 meters in front of the vehicle. The track followed by the vehicle is shown in Figure \ref{fig:tracks}. In this experiment, we generated only local goals based on the description given in the Navigtion stack (Section \ref{sec:nav}). 

\begin{figure}[h]
    \includegraphics[width = 8.7 cm, height = 7.5 cm]{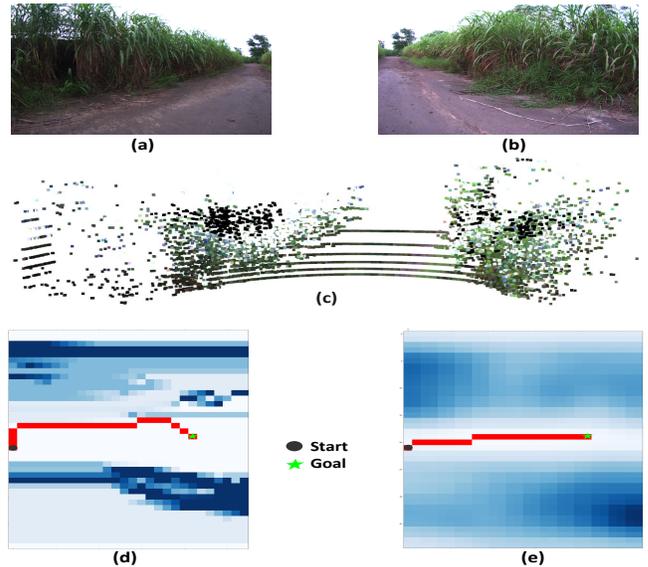}
    \caption{Real world implementation: (a) and (b) are the images from left and right camera. (c) Images projected on the VLP-16 point cloud data. (d) and (e) shows a comparison between the path generated from (d) manually handcrafted cost-map and (e) \camel ~predicted cost-map.}
    \label{fig:cost_comp}
    \vspace{-0.5cm}
\end{figure}

Moving from simulation to real-world implementation, the vehicle performed robust maneuvers with modification only for maximum throttle and steering constraints. This showcases the ability of \camel ~for direct simulation to real world (Sim2Real) application. Figure \ref{fig:cost_comp} shows an instance from the ground testing in Figure \ref{fig:tracks}(b). An imprecise calibration for the yaw angle of camera resulted in a complete failure of handcrafted cost-map \ref{fig:cost_comp}(d). However, the learned cost-map as shown in Figure \ref{fig:cost_comp}(e) is equivalent to the cost-map generated from calibrated data. \camel ~was able to differentiate the features despite perturbation in the system and predict a close to optimal path. 

While navigating we observed at instances where Offseg gives subpar segmentation, the cost-map generated maintains its prediction quality suggesting well balanced weight distribution to geometric information from LiDAR during training. During experiments vehicle opted to traverse through grass patch over puddle indicating in authors view, the involvement of LiDAR intensity for vegetation prediction but more exhaustive study is required to ascertain the observation.

Unlike HDL-64E LiDAR in the simulation, the VLP-16 gives 4$\times$ sparser point cloud. The framework still generates a cost-map that has acceptable throttle and steering commands. The algorithm could also classify negative obstacle from a distance assigning high grid costs hence steering away from the location.  
The inference speed of the whole framework with two Offseg predictions, a cost-map prediction and the navigation stack was tested on Jetson AGX Xavier. The Offseg with BisenetV2\cite{yu2020bisenet} gave an inference speed of 92 ms for each image whereas the FCN model has 4.5 ms inference speed. The whole data processing and navigation stack took on an average 0.95 seconds to process. This computation time can be reduced by multi-processing and optimization. 

\section{Conclusion and Future Work}
In this work, we presented \camel~ a FCN based deep learning framework to navigate a UGV reliably in outdoor environments. Fusing RGB images and point cloud data we learn a cost-map from expert human demonstrations completely eliminating a handcrafted cost-map. Our approach is validated in simulation and real-world off-road terrains showcasing highly robust and adaptive motion by the UGV. The framework exhibited invariancy to perturbations in sensor calibrations and system biases making manually crafted cost-maps completely obsolete.

Future work will target extensive study of LiDAR intensity and learning vegetation features in different climatic conditions. Another interesting direction is the effective sensing of negative obstacles such as steep troughs which is difficult to classify due to low LiDAR reflectance. 
\bibliographystyle{IEEEtran}
\bibliography{Reference}

\end{document}